\pdfoutput=1

\documentclass[11pt]{article}

\usepackage[]{acl}

\usepackage{float}
\usepackage{times}
\usepackage{latexsym}
\usepackage{tabularx}         
\usepackage{booktabs}         
\usepackage{multirow,multicol}         
\usepackage{graphicx}
\usepackage{cuted}

\usepackage[T1]{fontenc}

\usepackage[utf8]{inputenc}

\usepackage{microtype}

\title{A Side-by-side Comparison of Transformers for \\English Implicit Discourse Relation Classification}

\newcommand*{\email}[1]{\texttt{#1}}
\renewcommand{\thefootnote}{\fnsymbol{footnote}}

\author{
Bruce W. Lee$^{1,2}$, \ 
BongSeok Yang$^{2}$, \ 
Jason Hyung-Jong Lee$^{2}$\ 
\\ 
$^{1}$University of Pennsylvania - PA, USA \\
$^{2}$LXPER AI Research - Seoul, South Korea \\
 \email{brucelws@seas.upenn.edu} \\
 \email{bongseok@lxper.com} \\
 \email{jasonlee@lxper.com} \\}

\begin{document}
\maketitle

\renewcommand*{\thefootnote}{\arabic{footnote}}

\begin{abstract}
Though discourse parsing can help multiple NLP fields, there has been no wide language model search done on implicit discourse relation classification. This hinders researchers from fully utilizing public-available models in discourse analysis. This work is a straightforward, fine-tuned discourse performance comparison of seven pre-trained language models. We use PDTB-3, a popular discourse relation annotated dataset. Through our model search, we raise SOTA to 0.671 ACC and obtain novel observations. Some are contrary to what has been reported before \citep{shi2019next}, that sentence-level pre-training objectives (NSP, SBO, SOP) generally fail to produce the best performing model for implicit discourse relation classification. Counterintuitively, similar-sized PLMs with MLM and full attention led to better performance. 
\end{abstract}

\section{Introduction}
\begin{table*}[h]
\centering
\resizebox{\textwidth}{!}{%
\begin{tabular}{lccccccc}
\cmidrule(lr){1-8}
\textbf{Configurations}    & \textbf{ALBERT}$_{large}$ & \textbf{BART}$_{large}$  & \textbf{BigBird-R.} & \textbf{DeBERTa}$_{large}$ & \textbf{Longformer}$_{large}$ & \textbf{RoBERTa}$_{large}$ & \textbf{SpanBERT}$_{large}$\\ 
\cmidrule(lr){1-1}\cmidrule(lr){2-2}\cmidrule(lr){3-3}\cmidrule(lr){4-4}\cmidrule(lr){5-5}\cmidrule(lr){6-6}\cmidrule(lr){7-7}\cmidrule(lr){8-8}
Release        & 2019        & 2020              & 2020          & 2020               & 2020          & 2019          & 2020\\
Parameters     & 17M         & 406M              & -             & 350M               & 435M          & 340M          & 340M\\
Hidden         & 1024        & 1024              & -             & 1024               & 1024          & 1024          & 1024    \\
Layers         & 24 (Enc)    & 24 (Enc+Dec)$^{*}$& -             & 24 (Enc)           & 24 (Enc)      & 24 (Enc)      & 24 (Enc)\\
Attention Heads& 16          & 16                & -             & 16                 & 16            & 16            & 16      \\
Self-Attention & Full        & Full              & Block-Sparse  & Full$^{**}$        & Global+Window & Full          & Full\\
Max Seq. Length& 512         & 512               & 4096          & 512                & 4096          & 512           & 512\\
Pre-train Obj. & MLM \& SOP  & TI \& SS          & -             & MLM                & MLM           & MLM           & MLM \& SBO\\
\cmidrule(lr){1-8}
\end{tabular}
}
\caption{\label{Table 1} Tested language models and their varying configurations. $^{*}$: BART follows the original encoder-decoder architecture, 12 layers allocated for each. $^{**}$: DeBERTa uses disentangled attention. MLM: masked language modelling. SOP: sentence order prediction. SBO: span boundary objective. TI: text infilling. SS: sentence shuffling.}
\vspace{-4mm}
\end{table*} 
An utterance has multiple dimensions of meaning. Discourse relation classification identifies one such dimension: the coherence relation between clauses or sentences arising from low-level textual cues \citep{zhao2022revisiting, webber2019penn}. This makes the task important to several NLP fields, including multi-party dialogue analysis \citep{li2022survey}, social media postings analysis \citep{siskou-etal-2022-automatized}, and student literary writing analysis \citep{fiacco-etal-2022-toward}. A discourse relation is often marked with explicit connectives such as \textit{but, because, and}. Consider the following example:

\begin{quote}
Although Philip Morris typically tries to defend the rights of smokers, [\textit{"this has nothing to do with cigarettes, nor will it ever," the spokesman says}]$_{Arg1}$. [\underline{But}]$_{Conn}$ [\textbf{some anti-smoking activists disagree}]$_{Arg2}$, expressing anger... $\rightarrow$ \texttt{Comparison.Contrast}
\end{quote}

The explicit connective, \underline{Conn} (But), is informative. Hence, it is fairly easy to know that the two arguments, \textit{Arg1} and \textbf{Arg2}, are compared, likely in a contrasting relationship rather than similarity. This task is often referred to as explicit discourse relation classification. \citet{pitler-nenkova-2009-using} achieves a 94.15\% accuracy (4-way) with Naive Bayes.

Implicit discourse relation classification, on the other hand, aims to classify discourse relationships in cases without an explicit connective. It has received constant attention \citep{li2022survey} since the release of Penn Discourse Tree Bank 2.0 (PDTB-2) \cite{prasad2008penn}. Consider the following:

\begin{quote}
["\textit{Last year we probably bought one out of every three new deals,}]$_{Arg1}$," he says. "[\textbf{This year, at best, it's in one in every five or six.}]$_{Arg2}$" $\rightarrow$ \texttt{Comparison.Contrast}
\end{quote}

Without an explicit connective, \underline{Conn}, discourse relation classification only relies on low-level semantic cues from the arguments, \textit{Arg1} and \textbf{Arg2}. Such "implicit" discourse relation classification is very challenging as it requires a language model to conceptualize the unstated goal the speaker is trying to achieve, not only the literal content \citep{shi2019next, sileo2019discourse}.

With XLNet$_{large}$ \citep{yang2019xlnet} achieving $\sim$60\% accuracy \citep{kim-etal-2020-implicit}, pre-trained language models showed promising improvements from the past studies: Maximum-Entropy Learning ($\sim$40\% F1) \citep{lin2014pdtb}, Adversarial Network ($\sim$46\% ACC) \citep{qin-etal-2017-adversarial}, Seq2Seq + Memory Network ($\sim$48\% ACC) \citep{shi-demberg-2019-learning}. Implicit discourse relation classification gives relatively small textual information for a language model to infer from. Thus, pre-training large text helps establish typical relations within/across clauses and sentences \citep{shi2019next}.

Pre-trained language models, like BERT \citep{devlin2018bert}, follow transformer-type \citep{vaswani2017attention} architecture and have only been recently introduced into implicit discourse relation classification \citep{kishimoto2020adapting}. To the best of our knowledge, BERT and XLNet are the only pre-trained language models (fine-tuned and) evaluated for implicit discourse relation classification on PDTB-3 \citep{kim-etal-2020-implicit}. However, language models vary in architecture, training objective, data, etc.

Instead of performing a focused study on a single model, we fine-tune seven state-of-the-art (SOTA) language models \textbf{(\S2)}. Our wider approach brings weaknesses \textbf{(\S5)} (as we ignore some model-specific characteristics), but it allows the bird's-eye view of several downstream performances in PDTB-3 \textbf{(\S3)} \citep{webber2019penn} and raises SOTA ($\sim$67\% ACC) on \citet{kim-etal-2020-implicit}'s evaluation protocol. By contrasting performances, we show that certain language model characteristics can benefit implicit discourse relation classification. 

Additionally, we take the best-performing language model and check if the "full-sentence(s)" setup gives better performance \textbf{(\S3.4)}. As we elaborate further in the following sections, our sanity checks on PDTB-3 hint that some argument annotations are questionable in terms of consistency and coverage. Hence, implicit discourse relation classification accuracy might improve by simply training the language model with a full sentence(s) instead of human-annotated argument spans (\textit{Arg1} and \textbf{Arg2}). We evaluate this idea toward the end. 

\section{Background}
\begin{table*}[h]
\centering
\resizebox{\textwidth}{!}{%
\begin{tabular}{lccccccc}
\cmidrule(lr){1-8}
& \textbf{ALBERT}$_{large}$ & \textbf{BART}$_{large}$  & \textbf{BigBird-R.} & \textbf{DeBERTa}$_{large}$ & \textbf{Longformer}$_{large}$ & \textbf{RoBERTa}$_{large}$ & \textbf{SpanBERT}$_{large}$\\ 
\cmidrule(lr){1-8}
\multicolumn{8}{c}{\textbf{Hyperparameters}}\\
\cmidrule(lr){1-8}
Learning Rate   &  5e-6	     & 5e-6	 & 5e-6	     & 2e-6	          & 5e-6	     & 2e-6	     & 5e-6\\
\cmidrule(lr){1-8}
\multicolumn{8}{c}{\textbf{a: Argument Spans}}\\
\cmidrule(lr){1-8}
Accuracy   & 0.565	     & 0.657	 & 0.649	     & 0.671	          & 0.668	     & 0.670	     & 0.627\\
Variance   & 2.53e-4     & 2.15e-4   & 4.02e-4       & 2.70e-4            & 2.15e-4      & 3.32e-4       & 1.78e-4\\
\cmidrule(lr){1-8}
\multicolumn{8}{c}{\textbf{b: Full Sentence(s)}}\\
\cmidrule(lr){1-8}
Accuracy   & 0.534	     & 0.629	 & 0.620	     & 0.634	          & 0.627	     & 0.617	     & 0.598\\
Variance   & 2.27e-4	 & 4.28e-4	 & 2.79e-4 	 & 3.75e-4	          & 4.18e-4	 & 3.62e-4	     & 2.84e-4\\
\cmidrule(lr){1-8}
\end{tabular}
}
\caption{\label{Table 2} Language model performances (test set) on Level-2 14-way implicit discourse relation classification.}
\vspace{-4mm}
\end{table*} 
The pre-train and fine-tune paradigm have been led by the remarkable downstream task performances of pre-trained language models \citep{kalyan2021ammus, devlin2018bert}. For several NLP tasks, a pre-trained language model could have likely done a fine job at learning syntax, semantics, and world knowledge -- given enough data and model size \citep{wang-etal-2019-tell}.

A pre-trained language model's competence in discourse was questionable until \citet{shi2019next} proposed that BERT's pre-training objective can benefit implicit discourse relation classification. However, \citet{iter2020pretraining} hints that BERT is not the language model best suited to the task. 

Implicit discourse relation classification is an active area of research \citep{kurfali2022contributions, zhao2022revisiting,kurfali2021probing, knaebel-2021-discopy, munir2021memorizing, kurfali2021let, kishimoto2020adapting, bourgonje2019explicit, shi2019next, bai2018deep, dai2018improving, rutherford2017systematic}. However, there has been no wide-range model study on implicit discourse relation classification, limiting a researcher's scope of model choice. This issue is further complicated by the fact that discourse task performances do not always correlate with popular semantics-based natural language understanding (NLU) scores, such as GLUE \citep{sileo2019discourse}. Thus, it is difficult to predict which language model can perform well without a dedicated empirical exploration.

With the a version update to Penn Discourse Tree Bank (PDTB-3) \citep{webber2019penn} and the correspondingly updated evaluation method \citep{kim-etal-2020-implicit}, we fine-tune seven language models to implicit discourse relation classification.

The chosen language models are: RoBERTa$_{large}$ \citep{liu2019roberta}, ALBERT$_{large}$ \citep{lan2019albert}, BigBird-RoBERTa$_{large}$ \citep{zaheer2020big}, BART$_{large}$ \citep{lewis2020bart}, Longformer$_{large}$ \citep{beltagy2020longformer}, SpanBERT$_{large}$ \citep{joshi2020spanbert}, DeBERTa$_{large}$ \citep{he2020deberta}. These models are selected with diversity in mind, especially in terms of input sequence length, attention type, and pre-train objectives. These models follow the popular transformer architecture \citep{vaswani2017attention}, and we will not review each model in detail. A brief comparison is shown in Table 1. 

\section{Experiments}
\subsection{Data Preparation}
We obtained the official PDTB-3 data from the Linguistic Data Consortium\footnote{www.ldc.upenn.edu}. PDTB-3 is a large-scale resource of annotated discourse relations and their arguments over the 1 million words Wall Street Journal Corpus \citep{marcus-etal-1993-building}. From a public repository\footnote{github.com/najoungkim/pdtb3}, we retrieved the corresponding evaluation script \citep{kim-etal-2020-implicit}. We describe some characteristics of the evaluation protocol below.

\textbf{Cross-validation} is used on the section level to preserve paragraph and document structures. Cross-validation likely solves label sparsity issue \citep{shi-demberg-2017-need}. The 25 sections of PDTB-3 are divided into 12 folds with 2 development, 2 test, and 21 training sections in each fold. The sliding window of two sections is used, creating 12 folds.

\textbf{Label set} is composed of 14 senses on L2 discourse relations (see Appendix B). Only the senses with $\geq$100 instances are used. This is to produce results that are in align with \citet{kim-etal-2020-implicit}. This alignment is crucial as we directly compared our results against fine-tuend BERT from \citet{kim-etal-2020-implicit}, which is trained with next sentence prediction (NSP) objective. Multiply-annotated labels become separate training instances. 
\subsection{Fine-Tuning}
To ensure reproducibility, we only take pre-trained language models from the now ubiquitous Huggingface \citep{wolf2019huggingface} {\fontfamily{qcr}\selectfont transformers} library. Fine-tuning was done with PyTorch \citep{NEURIPS2019_9015} and our scripts are publicly available.

During fine-tuning, each training instance is a concatenation of two arguments ($=$ sequence of tokens in \textit{Arg1} and \textbf{Arg2}). BERT-type models carry special tokens ([CLS], [SEP], [EOS]) for segmentation: [CLS], \textit{Arg1}$_{1}$ ... \textit{Arg1}$_{N}$, [SEP], \textbf{Arg2}$_{1}$ ... \textbf{Arg2}$_{M}$, [EOS]. Depending on the model, these special tokens are modified or completely removed.

As for hyperparameter searches, we mostly focus on the learning rate. We use the popular AdamW optimizer with a linear scheduler (no warm-up steps). As for the learning rate, we start from 2e-5, a value commonly used for text classification since \citet{sun2019fine}. We test lower learning rates of 2e-6 and 5e-6; we find that 5e-6 (which is slightly lower than what is usually used in sequence classification) performs best for almost all models. The batch size is 8 and the max input length is set at 256.

Lastly, for each experiment step (i.e. BART on fold 1), we train for 10 epochs with an early stop. The training stops if the current epoch's validation loss (see development set \textbf{\S3.1}) did not decrease from the previous epoch. Model training time, GPU, language model repository address, and other details on hyperparameters are in Appendix C. 
\subsection{Evaluation and Observations}
\begin{table*}[h]
\vspace{-4mm}
\centering
\resizebox{\textwidth}{!}{%
\begin{tabular}{lccccccc}
\cmidrule(lr){1-8}
 & \textbf{ALBERT}$_{large}$ & \textbf{BART}$_{large}$  & \textbf{BigBird-R.} & \textbf{DeBERTa}$_{large}$ & \textbf{Longformer}$_{large}$ & \textbf{RoBERTa}$_{large}$ & \textbf{SpanBERT}$_{large}$\\ 
\cmidrule(lr){1-8}
\multicolumn{8}{c}{\textbf{a: Argument Spans}}\\
\cmidrule(lr){1-8}
Accuracy   & 0.566	     & 0.663	 & 0.653	     & 0.673	          & 0.669	     & 0.670	     & 0.629\\
Variance   & 2.94e-4     & 1.33e-4   & 1.62e-4       & 2.47e-4            & 1.68e-4      & 1.01e-4       & 2.03e-4\\
\cmidrule(lr){1-8}
\multicolumn{8}{c}{\textbf{b: Full Sentence(s)}}\\
\cmidrule(lr){1-8}
Accuracy   & 0.567	     & 0.660	 & 0.645	     & 0.656	          & 0.661	     & 0.652	     & 0.639\\
Variance   & 3.92e-4	 & 4.59e-4	 & 3.50e-4 	 & 1.85e-4	          & 2.56e-4	 & 4.10e-4	     & 3.45e-4\\
\cmidrule(lr){1-8}
\end{tabular}
}
\caption{\label{Table 3} Language model performances (\textbf{dev set}) on Level-2 14-way implicit discourse relation classification.}
\vspace{-4mm}
\end{table*} 
In Table 2-a, we report the mean test set accuracy of 12 folds along with variance. This is in alignment with what was recommended by \citet{kim-etal-2020-implicit}. Development set performances are given in Table 3 to facilitate reproducibility. For multiply-annotated labels (also discussed in \textbf{\S3.1}), the model only has to get one label correct. We reach some surprising observations, which we share below.

\textbf{1) Sentence-level pre-train objectives are not necessary to create best-performing models.} This is contrary to \citet{shi2019next}, which proposed that NSP helps implicit discourse relation classification after conducting an ablation study on BERT. Their finding was intuitive as well because implicit discourse relation classification aims to find the relationship between two argument spans. 

But in a more general scope, the necessity of NSP has been questioned multiple times \citep{yang2019xlnet,lample2019cross}. In other words, NSP -- or any other sentence-level pre-train objective for that matter -- could have been only helpful in some specific ablation study of BERT-type models but not in other cases \citep{liu2019roberta}. We obtain supporting results in Table 2-a, where language models with sentence-level objectives performed worse than MLM-only models given similar model sizes (ALBERT is an exception). 

\textbf{2) Long-document modifications (mostly done by altering attention schemes of an existing model) decrease the original model performance.} At first, we postulated that long-document models could lead to performance increases because they can learn long-span discourse relations during pre-training. But using sparse or block attention mechanisms eventually led to a performance decrease. 

The decrease is clearly demonstrated by BigBird-RoBERTa$_{large}$ and Longformer$_{large}$. Both models start from the existing RoBERTa$_{large}$ checkpoint and modify it to process longer sequences. Such modifications achieved performance increases in other NLP tasks like question-answering, coreference resolution, and some cases of sequence classification. But implicit discourse relation classification, which requires the model's understanding of dense discourse relations hidden within a few tokens, long-document modification is a drawback.

\textbf{3) The simplest combination of MLM and full attention is best suited for implicit discourse relation classification.} We are making this argument within the scope of what we have tested. We believe that MLM and full attention (e.g., RoBERTa, DeBERTa) work best because the model has to make inferences based on a relatively small number of tokens. Hence, trivial textual cues should not be risked being overlooked. MLM, with full attention, forces every token to attend to every other and learn the token-specific relations, likely to lose the least textual cues and nuances. 

\subsection{Train Full Sentence or Argument Span?}
Following the aforementioned observations, we postulated that fine-tuning language models using full sentence(s) could further improve classification accuracy. By full sentence(s), we refer to the sentence(s) (usually up to two) that the annotated argument spans appeared. We had two reasons for our postulation: 1. textual cues that hint at underlying discourse relation could be spread throughout the sentence(s), 2. argument span annotation is sometimes inconsistent, especially at punctuation marks, unnecessarily confusing the language model. Implicit discourse relation classification has rarely been tested using the full sentence.
  
We built an argument matcher to find the source sentence of each annotated argument span. For inter-sentential relations, we only considered argument spans that came from two adjacent source sentences. We share the test set results in Table 2-b. The results bring us to our fourth observation.

\textbf{4) As input, concatenating argument spans generally perform better than full sentence(s).} Opposed to our postulation, using full sentence(s) as input decreased performance on the test set. Though we see mixed results on the development set in Table 3, training full sentences as input generally decrease performance. But when it comes to implicit discourse sense classification from the raw text (that means in practical, end-to-end applications), the benefits of using argument spans must be weighed against the low accuracies (50\% $\sim$ 60\%) of the available argument extractors.  

\section{Conclusion}
Researchers often build or modify a neural network to improve task performance. While such effort is essential, this paper shows that SOTA can also be raised through extensive search and application of existing resources. Through a side-by-side comparison of seven PLMs, we also make handy observations on pre-training objectives, long-document modifications, and full-sentence setups. Though some might consider these phenomena rather expected, nothing is scientifically conclusive until an analysis is performed at an adequate scale. We hope that our report helps researchers working towards discourse understanding, and we continue to discuss the missing details in the appendices.

\section{Acknowledgement}
We thank the anonymous reviewers for their crisp and realistic advices on cleaning the language of the paper and experiments. Most of the review opinions were accepted and were reflected in the paper as they were valid. 

\bibliography{anthology,custom}
\bibliographystyle{acl_natbib}

\appendix

\section{``Full sentence(s)'' Experiment}
\subsection{What Makes the Experiment Important?}
This section is a continuation of \S3.4. Here, we discuss implicit discourse relation classification from raw sentence(s), which we believe is the best practical example of real-world applications of the related fields. Such an \textit{end-to-end} concept has been popularized through CoNLL-2016 \citep{xue2016conll} and CoNLL-2015 \citep{xue2015conll}, and most systems develop a separate argument span identification model. Then, the identified argument spans would be fed to the discourse relation classification model for sense prediction (refer to examples given in \textbf{\S1}) \citep{he-etal-2020-transs}. 

Such a double-step process makes sense. Indeed, feeding the exact argument spans (that only contain the tokens that imply a certain discourse sense) will increase sense prediction performance.

But the problem arises because identifying argument spans from raw sentence(s) is a low accuracy operation \citep{knaebel-2021-discopy}. A wrong span identification eventually leads to error propagation, providing a discourse relation classification model that lacks textual information. We give a theoretical error propagation example and conduct a simple experiment to prove our point. 
\subsection{Theoretical Example of Error Propagation}
-----

\noindent 1. A set of two raw sentences is given. 
\begin{itemize}
\item[] "Last year we probably bought one out of every three new deals," he says. "This year, at best, it's in one in every five or six."
\end{itemize}

-----

\noindent 2. Where correct argument spans are as below. 
\begin{itemize}
\item[] ["\textit{Last year we probably bought one out of every three new deals,}]$_{Arg1}$" he says. "[\textbf{This year, at best, it's in one in every five or six.}]$_{Arg2}$"
\end{itemize}

-----

\noindent 3. But an argument span identification model often makes wrong predictions (best system \citep{oepen2016opt} at CoNLL-2016 scores 52.02 F1, for exact span match). 
\begin{itemize}
\item[] ["\textit{Last year we probably bought one}]$_{Arg1}$ out of every three new deals," he says. "This year, at best, [\textbf{it's in one in every five or six.}]$_{Arg2}$"
\end{itemize}

-----

\noindent 4. Now, compare the amount of textual information passed over to the implicit discourse relation classification model, under three setups. Note that setup 1 cannot be used in real-world settings because it requires PDTB-3's gold annotations.

\textbf{Setup 1)} PDTB-3 (with gold annotations)
\begin{itemize}
\item[] Last year we probably bought one out of every three new deals This year, at best, it's in one in every five or six.
\end{itemize}

\textbf{Setup 2)} A low accuracy argument span model
\begin{itemize}
\item[] Last year we probably bought one it's in one in every five or six.
\end{itemize}

\textbf{Setup 3)} Full sentence(s)
\begin{itemize}
\item[] "Last year we probably bought one out of every three new deals," he says. "This year, at best, it's in one in every five or six."
\end{itemize} 
\subsection{Experiment on Error Propagation}
\begin{table}[t]

\begin{center}
\footnotesize

\begin{tabular}{l 
c @{\hspace{0.8ex}} c @{\hspace{0.8ex}} c @{\hspace{0.8ex}}}

\toprule
\multirow{2.4}{*}{\textbf{Fine-tuned PLM}}& 
\multicolumn{2}{c}{Argument Span}\\ 

\cmidrule(lr){2-3}

& ACC &F1\\ 
			
\midrule
BERT$_{large}$     
&0.912&0.742\\

\bottomrule
\end{tabular}
\end{center}
\caption{BERT's performance (12-folds test set) on PDTB-3's argument spans.}
\vspace{-4mm}
\end{table} 
Though not all tokens are valuable under a full sentence(s) setup, we can notice that it is a foolproof way to input all meaningful tokens. Table 4 reports the classification performance of BERT$_{large}$, which was trained to identify argument spans using PDTB-3. Our argument span scoring scheme approximately matches CoNLL-16's partial scoring scheme, essentially a relaxed version of conlleval. That means we consider a prediction correct if more than 70\% of argument span tokens are identified. For implicit discourse relation classification, a sense prediction is correct if it matches any of the multiply-annotated senses.

BERT's 0.912 ACC score implies that the model could correctly identify at least 70\% of the gold argument span tokens more than 9 out of 10 times. Nonetheless, error propagation detrimentally affected implicit discourse relation classification performance in Table 5. This empirically proves our ideas in Appendix A.1. 

\begin{table}[ht]
\begin{center}
\footnotesize

\begin{tabular}{l 
c @{\hspace{0.8ex}} c @{\hspace{0.8ex}} c @{\hspace{0.8ex}}}

\toprule
\multirow{2.4}{*}{\textbf{Fine-tuned PLM}}& 
\multicolumn{2}{c}{Implicit Sense}\\ 

\cmidrule(lr){2-3}
& ACC &F1  \\ 
			
\midrule
DeBERTa$_{large}$
&0.670&0.671 \\
\quad \quad \textit{with error propagation}    
&0.476&0.491\\
\quad \quad \textit{full sentence(s)}    
&0.634&0.637\\

\bottomrule
\end{tabular}
\end{center}
\caption{DeBERTa performances (12-fold test set) on PDTB-3's Level-2 14-way implicit discourse relation classification, but under three different pipeline setups.}
\end{table} 
\section{14-way Label Set}
\begin{table}[t]
\vspace{-4mm}
\begin{center}
\footnotesize
\begin{tabular}{lc}
\toprule
\textbf{Label}    & \textbf{Counts}\\ 

\midrule
Comparison.Concession     & 1494\\
Comparison.Contrast       & 983\\
Contingency.Cause         & 5785\\
Contingency.Cause+Belief  & 202\\
Contingency.Condition     & 199\\
Contingency.Purpose       & 1373\\
Expansion.Conjunction     & 4386\\
Expansion.Equivalence     & 336\\
Expansion.Instantiation   & 1533\\
Expansion.Level-of-detail & 3361\\
Expansion.Manner          & 739\\
Expansion.Substitution    & 450\\
Temporal.Asynchronous     & 1289\\
Temporal.Synchronous      & 539\\
\bottomrule

\end{tabular}
\caption{\label{Table 3} Counts of 14-way implicit discourse senses.}
\vspace{-4mm}
\end{center}
\end{table} 
\section{More on Fine-tuning Set Up}
We ran all our experiments on a single NVIDIA Tesla V100 GPU. Model train time and repositories are listed below. Training times below suppose no early stop. The performances reported in Table 2 are obtained \textbf{with} early stop.

\noindent \textbf{ALBERT}$_{large}$ 

- huggingface.co/albert-large-v1

- $\sim$2.4 days, for 12 folds $\times$ 10 epochs 

\noindent \textbf{BART}$_{large}$  

- huggingface.co/facebook/bart-large

- $\sim$3.6 days, for 12 folds $\times$ 10 epochs 

\noindent \textbf{BigBird-RoBERTa}$_{large}$ 

- huggingface.co/google/bigbird-roberta-large

- $\sim$3.2 days, for 12 folds $\times$ 10 epochs 

\noindent \textbf{DeBERTa}$_{large}$ 

- huggingface.co/microsoft/deberta-large

- $\sim$4.6 days, for 12 folds $\times$ 10 epochs 

\noindent \textbf{Longformer}$_{large}$ 

- huggingface.co/allenai/longformer-large-4096

- $\sim$11 days, for 12 folds $\times$ 10 epochs 

\noindent \textbf{RoBERTa}$_{large}$ 

- huggingface.co/roberta-large

- $\sim$2.9 days, for 12 folds $\times$ 10 epochs 

\noindent \textbf{SpanBERT}$_{large}$

- .../SpanBERT/spanbert-large-cased

- $\sim$2.9 days, for 12 folds $\times$ 10 epochs 
\end{document}